\documentclass[10pt,twocolumn,letterpaper]{article}

\usepackage{cvpr}
\usepackage{times}
\usepackage{epsfig}
\usepackage{graphicx}
\usepackage{amsmath}
\usepackage{amssymb}

\usepackage[pagebackref=true,breaklinks=true,letterpaper=true,colorlinks,bookmarks=false]{hyperref}

\usepackage{booktabs}
\usepackage{subcaption}
\usepackage{multirow}
\usepackage{mmstyle}
\newcommand*{\Scale}[2][4]{\scalebox{#1}{$#2$}}%

\newcommand{\algname}{GA-RPN}

\cvprfinalcopy

\newcommand{\cavan}[1]{{\color{blue}(cavan: {#1})}} 

\def\cvprPaperID{1843} 

\ifcvprfinal\fi
\begin{document}

\title{Region Proposal by Guided Anchoring}

\author{Jiaqi Wang$^{1}$\thanks{Equal contribution.} \quad Kai Chen$^{1*}$ \quad Shuo Yang$^2$ \quad Chen Change Loy$^3$ \quad Dahua Lin$^1$\\
	$^1$CUHK - SenseTime Joint Lab, The Chinese University of Hong Kong \\
	$^2$Amazon Rekognition \hspace{10pt} $^3$Nanyang Technological University\\
	{\tt\small \{wj017,ck015,dhlin\}@ie.cuhk.edu.hk}\hspace{10pt}
	{\tt\small shuoy@amazon.com}\hspace{10pt}
	{\tt\small ccloy@ntu.edu.sg}
}

\maketitle
\thispagestyle{empty}


\begin{abstract}

Region anchors are the cornerstone of modern object detection techniques.
State-of-the-art detectors mostly rely on a dense anchoring scheme,
where anchors are sampled uniformly over the spatial domain with
a predefined set of scales and aspect ratios.
In this paper, we revisit this foundational stage. Our study shows that
it can be done much more effectively and efficiently.
Specifically, we present an alternative scheme, named Guided Anchoring,
which leverages semantic features to guide the anchoring.
The proposed method jointly predicts the locations where the center of objects of interest
are likely to exist as well as the scales and aspect ratios at different locations.
On top of predicted anchor shapes, we mitigate the feature inconsistency with a feature adaption module.
We also study the use of high-quality proposals to improve detection performance.
The anchoring scheme can be seamlessly integrated into proposal methods and detectors.
With Guided Anchoring, we achieve $9.1\%$ higher recall on MS COCO
with $90\%$ fewer anchors than the RPN baseline.
We also adopt Guided Anchoring in Fast R-CNN, Faster R-CNN and RetinaNet,
respectively improving the detection mAP by $2.2\%$, $2.7\%$ and $1.2\%$.
Code will be available at~\url{https://github.com/open-mmlab/mmdetection}.

\end{abstract}


\section{Introduction}
\label{sec:intro}

Anchors are regression references and classification candidates to predict proposals
(for two-stage detectors) or final bounding boxes (for single-stage detectors).
Modern object detection pipelines usually begin with a large set of densely distributed anchors.
Take Faster RCNN~\cite{ren2015faster}, a popular object detection framework, for instance,
it first generates region proposals from a dense set of anchors and then
classifies them into specific classes and refines their locations via
bounding box regression.

There are two general rules for a reasonable anchor design: \emph{alignment} and \emph{consistency}.
Firstly, to use convolutional features as anchor representations, anchor centers
need to be well aligned with feature map pixels.
Secondly, the receptive field and semantic scope should be consistent with the scale and shape of anchors on different locations of a feature map.
The sliding window is a simple and widely adopted anchoring scheme following the rules.
For most detection methods, the anchors are defined by such a \emph{uniform}
scheme, where every location in a feature map is associated with $k$ anchors
with predefined scales and aspect ratios.

Anchor-based detection pipelines have been shown effective in both
benchmarks~\cite{everingham2015pascal,lin2014microsoft,geiger2012we,deng2009imagenet}
and real-world systems.
However, the uniform anchoring scheme described above is not necessarily
the optimal way to prepare the anchors.
This scheme can lead to two difficulties:
(1) A neat set of anchors of fixed aspect ratios has to be predefined for different problems. A wrong design may hamper the speed and accuracy of the detector.
(2) To maintain a sufficiently high recall for proposals, a large number of
anchors are needed, while most of them correspond to \emph{false} candidates that are irrelevant to the object of interests.
Meanwhile, a large number of anchors can lead to significant computational cost especially
when the pipeline involves a heavy classifier in the proposal stage.

In this work, we present a more effective method to prepare anchors,
with the aim to mitigate the issues of hand-picked priors.
Our method is motivated by the observation that objects are not distributed
evenly over the image.
The scale of an object is also closely related to the imagery content,
its location and geometry of the scene.
Following this intuition, our method generates sparse anchors in two steps:
first identifying sub-regions that may contain objects and then
determining the shapes at different locations.

Learnable anchor shapes are promising, but it breaks the aforementioned rule of consistency, thus presents a new challenge for learning anchor representations for accurate classification and regression.
Scales and aspect ratios of anchors are now variable instead of fixed,
so different feature map pixels have to learn adaptive representations that
fit the corresponding anchors.
To solve this problem, we introduce an effective module to adapt
the features based on anchor geometry.

We formulate a Guided Anchoring Region Proposal Network (\algname) with the
aforementioned guided anchoring and feature adaptation scheme.
Thanks to the dynamically predicted anchors, our approach achieves 9.1\%
higher recall with 90\% substantially fewer anchors than the RPN baseline
that adopts dense anchoring scheme.
By predicting the scales and aspect ratios instead of fixing them based on
a predefined list, our scheme handles tall or wide objects more effectively.
Besides region proposals, the guided anchoring scheme can be easily integrated
into any detectors that depend on anchors. Consistent performance gains can be achieved with our scheme.
For instance, GA-Fast-RCNN, GA-Faster-RCNN and GA-RetinaNet improve overall mAP by 2.2\%, 2.7\% and 1.2\%
respectively on COCO dataset over their baselines with sliding window anchoring.
Furthermore, we explore the use of high-quality proposals, and propose
a fine-tuning schedule using GA-RPN proposals, which can improve the performance
of any trained models, \eg, it improves a fully converged Faster R-CNN model
from 37.4\% to 39.6\%, in only $3$ epochs.

The main contributions of this work lie in several aspects.
(1) We propose a new anchoring scheme with the ability to predict non-uniform and
arbitrary shaped anchors other than dense and predefined ones.
(2) We formulate the joint anchor distribution with two factorized conditional
distributions, and design two modules to model them respectively.
(3) We study the importance of aligning features with the corresponding anchors
and design a feature adaption module to refine features based on the underlying
anchor shapes.
(4) We investigate the use of high-quality proposals for two-stage detectors
and propose a scheme to improve the performance of trained models.

\section{Related Work}
\label{sec:related}


\noindent
\textbf{Sliding window anchors in object detection.}
Generating anchors with the sliding window manner in feature maps has been
widely adopted by anchor-based various detectors.
The two-stage approach has been the leading paradigm in the modern era of
object detection.
Faster R-CNN~\cite{ren2015faster} proposes the Region Proposal Network (RPN) to 
generates object proposals. It uses a small fully convolutional network to map
each sliding window anchor to a low-dimensional feature. This design is also
adopted in later two-stage methods~\cite{dai2016r,lin2017feature,he2017mask}.
MetaAnchor~\cite{yang18metaanchor} introduces meta-learning to anchor generation.
There have been attempts~\cite{ghodrati2015deepproposal,gidaris2015object,najibi2016g,yang2016craft,zhang2017single,zhong2017cascade,cai18cascadercnn,chen2019hybrid}
that apply cascade architecture to reject easy samples at early layers or stages,
and regress bounding boxes iteratively for progressive refinement.
Compared to two-stage approaches, the single-stage pipeline skips object proposal generation and predicts bounding boxes and class scores in one evaluation.
Although the proposal step is omitted, single-stage methods still use anchor boxes produced by the sliding window.
For instance, SSD~\cite{liu2016ssd} and DenseBox~\cite{huang2015densebox}
generate anchors densely from feature maps and evaluate them like a multi-class RPN.
RetinaNet~\cite{lin2017focal} introduces focal loss to address the
foreground-background class imbalance. 
YOLOv2\cite{redmon2017yolo9000}
adopt sliding window anchors for classification and spatial location prediction so as to achieve a higher recall than its precedent.

\noindent
\textbf{Comparison and difference.}
We summarize the differences between the proposed method and conventional methods as follows.
(i) Primarily, previous methods (single-stage, two-stage and multi-stage) still
rely on dense and uniform anchors by sliding window. We discard the sliding
window scheme and propose a better counterpart to guide the anchoring and
generate sparse anchors, which has not been explored before.
(ii) Cascade detectors adopt more than one stage to refine detection bounding
boxes progressively, which usually leads to more model parameters and a decrease in inference speed.
These methods adopt RoI Pooling or RoI Align to extract aligned features for bounding boxes,
which is too expensive for proposal generation or single-stage detectors.
(iii) Anchor-free methods~\cite{huang2015densebox,jie2016scale,redmon2016you}
usually have simple pipelines and produce final detection results within a single stage.
Due to the absence of anchors and further anchor-based refinement,
they lack the ability to deal with complex scenes and cases.
Our focus is the sparse and non-uniform anchoring scheme and use of high-quality
proposals to boost the detection performance.
Towards this goal, we have to solve the misalignment and inconsistency issues which are specific to anchor-based methods.
(iv) Some single-shot detectors~\cite{zhang2017single,wu2018singleshot} refine
anchors by multiple regression and classification.
Our method differs from them significantly.
We do not refine anchors progressively, instead, we predict the distribution of
anchors, which is factorized as locations and shapes.
Conventional methods fail to consider the alignment between anchors and features
so they regress anchors (represented by $[x,y,w,h]$) for multiple times and breaks the alignment as well as consistency.
On the contrary, we emphasize the importance of the two rules,
so we only predict anchor shapes but fix anchor centers and adapt features based on the predicted shapes.


\begin{figure*}[t]
	\centering
	\includegraphics[width=0.9\linewidth]{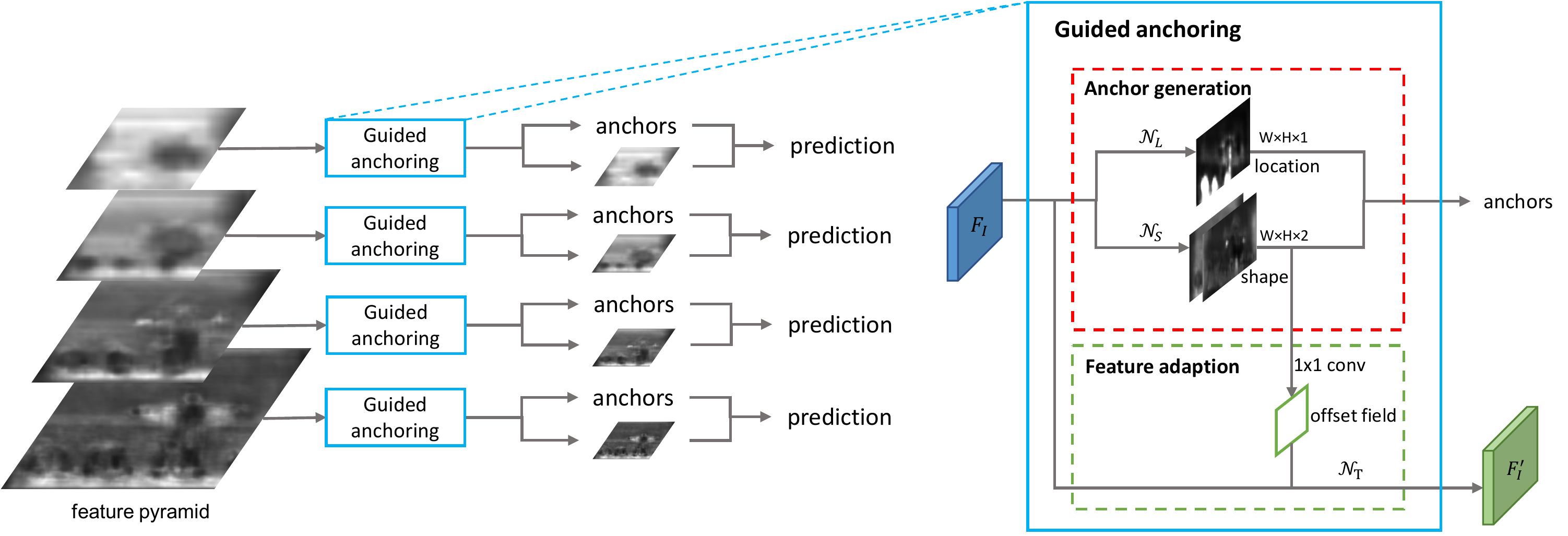}
	\caption{\small{An illustration of our framework. For each output feature map in the
		feature pyramid, we use an anchor generation module with two branches to
		predict the anchor location and shape, respectively. Then a feature
		adaption module is applied to the original feature map to make the
		new feature map aware of anchor shapes.}}
	\label{fig:framework}
	\vspace{-15pt}
\end{figure*}

\section{Guided Anchoring}
\label{sec:methodology}


Anchors are the basis in modern object detection pipelines. Mainstream
frameworks, including two-stage and single-stage methods,
mostly rely on a \emph{uniform} arrangement of anchors.
Specifically, a set of anchors with predefined scales and aspect ratios
will be deployed over a feature map of size $W \times H$, with a stride of
$s$. This scheme is inefficient, as many of the anchors are placed in
regions where the objects of interest are unlikely to exist.
In addition, such hand-picked priors unrealistically assume a set of fixed shape
(\ie,~scale and aspect ratio) for objects.


In this work, we aim to develop a more efficient anchoring scheme to arrange the anchors with learnable shapes,
considering the non-uniform distribution of objects' locations and shapes.
The guided anchoring scheme works as follows.
The location and the shape of an object can be characterized by a 4-tuple
in the form of $(x, y, w, h)$,
where $(x, y)$ is the spatial coordinate of the center, $w$ the width,
and $h$ the height. Suppose we draw an object from a given image $I$, then
its location and shape can be considered to follow a distribution conditioned
on $I$, as follows:
\begin{equation}
	p(x,y,w,h|I)=p(x,y|I)p(w,h|x,y,I).
\end{equation}
This factorization captures two important intuitions:
(1) given an image, objects may only exist in certain regions; and
(2) the shape, \ie, scale and aspect ratio, of an object closely relates to
its location.


Following this formulation, we devise an anchor generation module
as shown in the red dashed box of Figure~\ref{fig:framework}. This module
is a network comprised of two branches for location
and shape prediction, respectively.
Given an image $I$, we first derive a feature map $F_I$. On top of $F_I$,
the \emph{location prediction} branch yields a probability map that indicates
the possible locations of the objects, while the \emph{shape prediction}
branch predicts location-dependent shapes.
Given the outputs from both branches, we generate a set of \emph{anchors} by
choosing the locations whose predicted probabilities are above a certain threshold
and the most probable shape at each of the chosen locations.
As the anchor shapes can vary, the features at different locations should
capture the visual content within different ranges.
With this taken into consideration, we further introduce a
\emph{feature adaptation} module, which adapts the feature according to the anchor shape.


The anchor generation process described above is based on a single feature map.
Recent advances in object detection~\cite{lin2017feature,lin2017focal} show
that it is often helpful to operate on multiple feature maps at different levels.
Hence, we develop a multi-level anchor generation scheme, which collects
anchors at multiple feature maps, following the FPN architecture~\cite{lin2017feature}.
Note that in our design, the anchor generation parameters are shared
across all involved feature levels thus the scheme is parameter-efficient.

\subsection{Anchor Location Prediction}
\label{subsec:anchor-loc}

As shown in Figure~\ref{fig:framework}, the \emph{anchor location prediction}
branch yields a probability map $p(\cdot|F_I)$ of the same size as the input
feature map $F_I$, where each entry $p(i, j | F_I)$ corresponds to the location
with coordinate $((i + \frac{1}{2})s, (j + \frac{1}{2})s)$ on $I$, where $s$
is stride of the feature map, \ie,~the distance between neighboring anchors.
The entry's value
indicates the probability of an object's center existing at that location.

In our formulation, the probability map $p(i, j | F_I)$ is predicted using a sub-network $\cN_L$.
This network applies a $1 \times 1$ convolution to the base feature map $F_I$
to obtain a map of objectness scores, which are then converted to
probability values via an element-wise sigmoid function.
While a deeper sub-network can make more accurate predictions, we found
empirically that a convolutional layer followed by a sigmoid transform strikes
a good balance between efficiency and accuracy.

Based on the resultant probability map, we then determine the active regions
where objects may possibly exist by selecting those locations whose
corresponding probability values are above a predefined threshold $\epsilon_L$.
This process can filter out 90\% of the regions while still
maintaining the same recall.
As illustrated in Figure~\ref{fig:visualization}(b), regions like sky and ocean
are excluded, while anchors concentrate densely around persons and surfboards.
Since there is no need to consider those excluded regions, we replace the ensuing
convolutional layers by \emph{masked convolution}~\cite{li2017not,song2018beyond}
for more efficient inference.

\subsection{Anchor Shape Prediction}
\label{subsec:anchor-shape}

After identifying the possible locations for objects, our next step is to determine
the shape of the object that may exist at each location.
This is accomplished by the \emph{anchor shape prediction}
branch, as shown in Figure~\ref{fig:framework}.
This branch is very different from conventional bounding box regression,
since it does not change the anchor positions and will not cause misalignment
between anchors and anchor features.
Concretely, given a feature map $F_I$, this branch will predict
the best shape $(w, h)$ for each location, \ie,~the shape that may lead to the highest coverage with the nearest ground-truth bounding box.

While our goal is to predict the values of the width $w$ and the height $h$,
we found empirically that directly predicting these two numbers is not stable,
due to their large range.
Instead, we adopt the following transformation:
\begin{equation} \label{eq:shape_t}
	w = \sigma\cdot s\cdot e^{dw}, \quad h = \sigma\cdot s\cdot e^{dh}.
\end{equation}
The shape prediction branch will output $dw$ and $dh$ , which will then be mapped
to $(w, h)$ as above, where $s$ is the stride and $\sigma$ is an empirical
scale factor ($\sigma=8$ in our experiments).
This nonlinear transformation projects the output space from approximate
$[0, 1000]$ to $[-1, 1]$, leading to an easier and stable learning target.
In our design, we use a sub-network $\cN_S$ for shape prediction, which
comprises a $1 \times 1$ convolutional layer that yields a two-channel map
that contains the values of $dw$ and $dh$, and an element-wise
transform layer that implements Eq.\eqref{eq:shape_t}.

Note that this design differs essentially from the conventional anchoring schemes
in that every location is associated with just one anchor of the dynamically predicted shape
instead of a set of anchors of predefined shapes.
Our experiments show that due to the close relations between locations and shapes,
our scheme can achieve much higher recall than the baseline scheme.
Since it allows arbitrary aspect ratios, our scheme can better capture
those extremely tall or wide objects.

\subsection{Anchor-Guided Feature Adaptation}
\label{subsec:feature-adaption}

In the conventional RPN or single stage detectors where the sliding window scheme
is adopted, anchors are uniform on the whole feature map, \ie, they share the
same shape and scale in each position. Thus the feature map can learn
consistent representation.
In our scheme, however, the shape of anchors varies across locations.
Under this condition, we find that it may not be a good choice to follow
the previous convention~\cite{ren2015faster}, in which a fully convolutional
classifier is applied uniformly over the feature map.
Ideally, the feature for a large anchor should encode the content over
a large region, while those for small anchors should have smaller scopes
accordingly.
Following this intuition, we further devise an
\emph{anchor-guided feature adaptation} component, which will transform
the feature at each individual location based on the underlying anchor shape,
as
\begin{equation}
	\vf_i^\prime = \cN_T(\vf_i, w_i, h_i),
\end{equation}
where $\vf_i$ is the feature at the $i$-th location, $(w_i, h_i)$ is the
corresponding anchor shape.
For such a location-dependent transformation, we adopt a
$3\times3$ deformable convolutional layer~\cite{dai16rfcn} to implement $\cN_T$.
As shown in Figure~\ref{fig:framework}, we first predict an offset
field from the output of anchor shape prediction branch, and then apply
deformable convolution to the original feature map with the offsets to
obtain $f_I^\prime$.
On top of the adapted features, we can then perform further
classification and bounding-box regression.

\subsection{Training}

\noindent
\textbf{Joint objective.}
The proposed framework is optimized in an end-to-end fashion using a multi-task loss.
Apart from the conventional classification loss $\cL_{cls}$ and regression loss $\cL_{reg}$,
we introduce two additional losses for the anchor localization $\cL_{loc}$ and
anchor shape prediction $\cL_{shape}$.
They are jointly optimized with the following loss.
\begin{equation}
	\cL = \lambda_1\cL_{loc} + \lambda_2\cL_{shape} + \cL_{cls} + \cL_{reg}.
\end{equation}

\noindent
\textbf{Anchor location targets.}
To train the anchor localization branch, for each image we need a binary label map where $1$ represents a valid location to place an anchor and $0$ otherwise.
In this work, we employ ground-truth bounding boxes for guiding the binary label map generation.
In particular, we wish to place more anchors around the vicinity of an object's center, while fewer of them far from the center.
Firstly, we map the ground-truth bounding box $(x_g, y_g, w_g, h_g)$ to the
corresponding feature map scale, and obtain $(x_g^\prime, y_g^\prime, w_g^\prime, h_g^\prime)$.
We denote $\cR(x, y, w, h)$ as the rectangular region whose center is
$(x, y)$ and the size of $w\times h$.
Anchors are expected to be placed close to the center of ground truth objects to obtain larger initial IoU, thus we define three types of regions for each box.

\noindent
(1) The center region $CR=\cR(x_g^\prime,y_g^\prime,\sigma_1w^\prime, \sigma_1h^\prime)$
defines the center area of the box. Pixels in $CR$ are assigned as positive samples.

\noindent
(2) The ignore region $IR=\cR(x_g^\prime,y_g^\prime,\sigma_2w^\prime, \sigma_2h^\prime)\backslash CR$
is a larger ($\sigma_2>\sigma_1$) region excluding $CR$. Pixels in $IR$ are marked as ``ignore'' and excluded during training.

\noindent
(3) The outside region $OR$ is the feature map excluding $CR$ and $IR$.
Pixels in $OR$ are regarded as negative samples.

\begin{figure}[t]
	\centering
	\includegraphics[width=\linewidth]{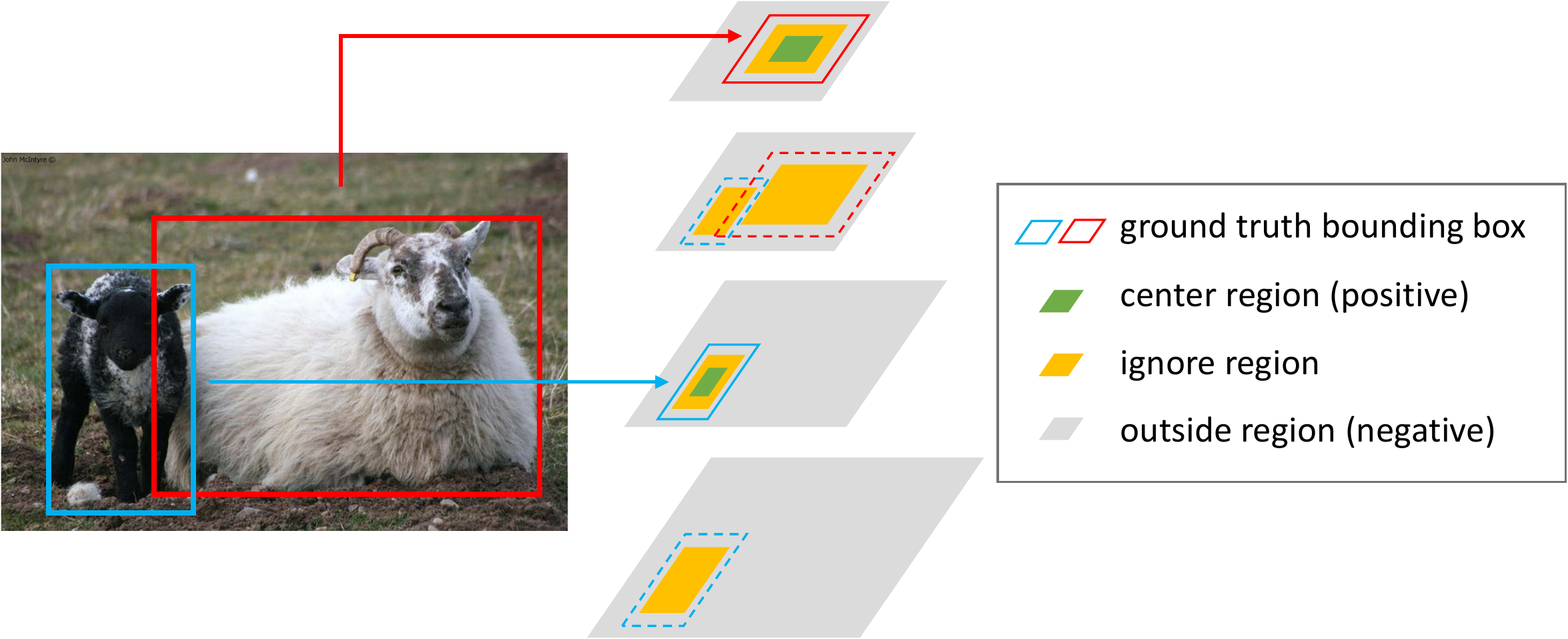}
	\caption{\small{Anchor location target for multi-level features. We assign ground
		truth objects to different feature levels according to their scales, and
		define $CR, IR$ and $OR$ respectively. (Best viewed in color.)}}
	\label{fig:location-target}
	\vspace{-15pt}
\end{figure}

Previous work~\cite{huang2015densebox} proposed the ``gray zone'' for balanced sampling,
which has a similar definition to our location targets but only works on a single feature map.
Since we use multiple feature levels from FPN, we also consider the influence
of adjacent feature maps.
Specifically, each level of feature map should only target objects of a specific scale range,
so we assign $CR$ on a feature map only if the feature map matches the scale
range of the targeted object.
The same regions of adjacent levels are set as $IR$, as shown in Figure~\ref{fig:location-target}.
When multiple objects overlap, $CR$ can suppress $IR$, and $IR$ can suppress $OR$.
Since $CR$ usually accounts for a small portion of the whole feature map, we use Focal Loss~\cite{lin2017focal} to train the location branch.

\noindent
\textbf{Anchor shape targets.}
There are two steps to determine the best shape target for each anchor.
First, we need to match the anchor to a ground-truth bounding box.
Next, we will predict the anchor's width and height which can
best cover the matched ground-truth.

Previous work~\cite{ren2015faster} assign a candidate anchor to the ground
truth bounding box that yields the largest IoU value with the anchor.
However, this process is not applicable in our case, since $w$ and $h$
of our anchors are not predefined but variables.
To overcome this problem, we define the IoU between a variable anchor
$a_{\textbf{wh}}=\{(x_0,y_0,w,h)|w>0,h>0\}$
and a ground truth bounding box $\text{gt}=(x_g, y_g, w_g, h_g)$ as follows,
denoted as vIoU.
\begin{equation} \label{eq:shape_approx}
	\text{vIoU}(a_{\textbf{wh}}, \text{gt})=\max_{w>0,h>0}\text{IoU}_{normal}(a_{wh}, \text{gt}),
\end{equation}
where $\text{IoU}_{normal}$ is the typical definition of IoU and $w$ and $h$
are variables.
Note that for an arbitrary anchor location $(x_0, y_0)$ and ground-truth gt,
the analytic expression of $\text{vIoU}(a_{\textbf{wh}}, \text{gt})$ is complicated,
and hard to be implemented efficiently in an end-to-end network.
Therefore we use an alternative way to approximate it. Given $(x_0, y_0)$,
we sample some common values of $w$ and $h$ to simulate the enumeration of
all $w$ and $h$. Then we calculate the IoU of these sampled anchors with gt,
and use the maximum as an approximation of $\text{vIoU}(a_{\textbf{wh}}, \text{gt})$.
In our experiments, we sample 9 pairs of $(w, h)$ to estimate $\text{vIoU}$
during training.
Specifically, we adopt the 9 pairs of different scales and aspect ratios used
in RetinaNet\cite{lin2017focal}.
Theoretically, the more pairs we sample, the more accurate the approximation is,
while the computational cost is heavier.
We adopt a variant of bounded iou loss~\cite{tychsen2018improving} to optimize
the shape prediction, without computing the target.
The loss is defined in Eq.~\eqref{eq:bounded-iou}, where $(w,h)$ and
$(w_{g},h_{g})$ denote the predicted anchor shape and the shape
of the corresponding ground-truth bounding box. $\cL_{1}$ is the smooth L1 loss.
\begin{equation}
\Scale[0.91]{\cL_{shape}=\cL_{1}(1-\min({\frac{w}{w_{g}}, \frac{w_{g}}{w}})) + \cL_{1}(1-\min({\frac{h}{h_{g}}, \frac{h_{g}}{h}})).}
\label{eq:bounded-iou}
\end{equation}

\begin{figure}[t]
	\centering
	\includegraphics[width=0.7\linewidth]{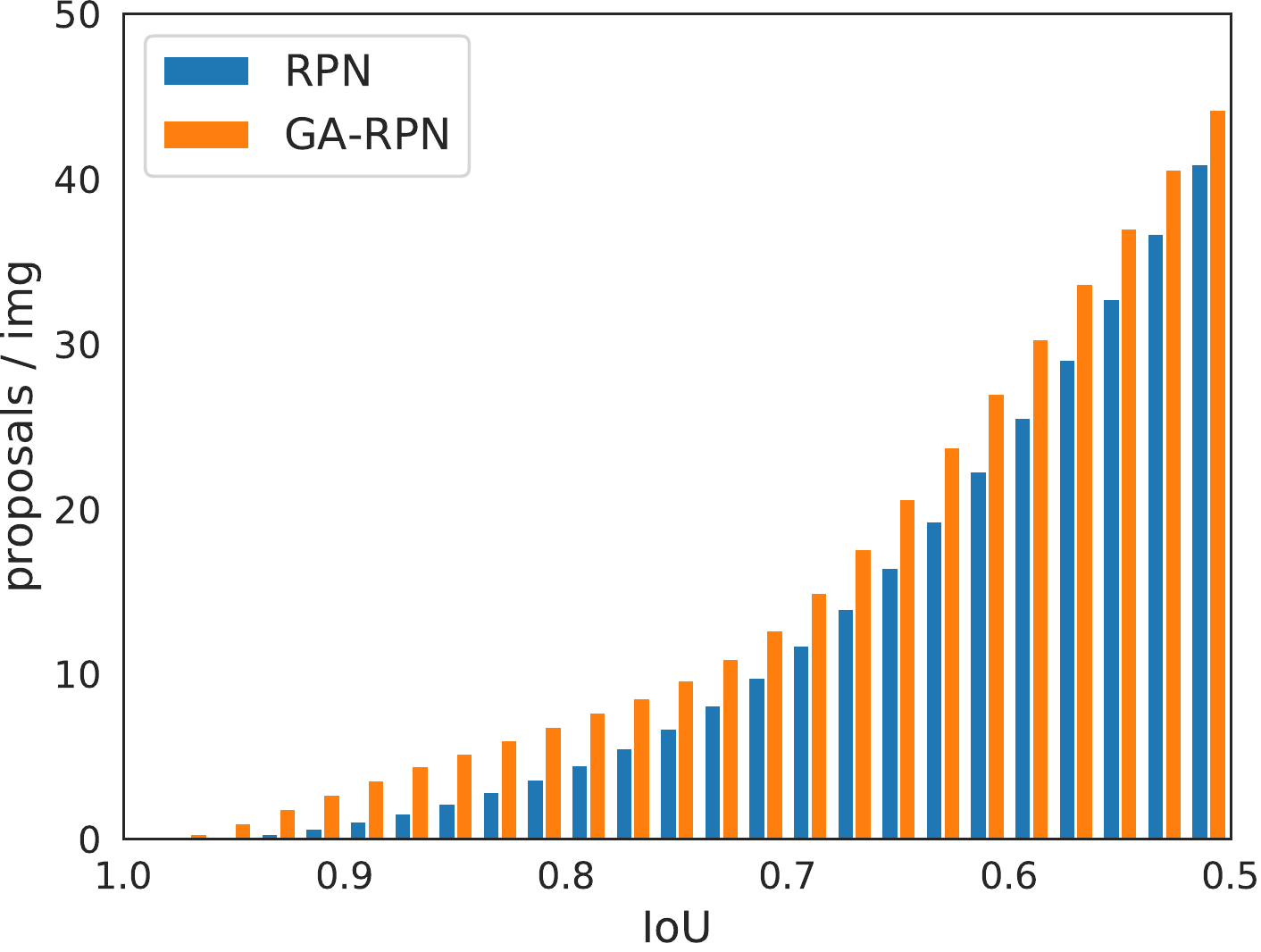}
	\caption{\small{IoU distribution of RPN and GA-RPN proposals. We show the
		accumulated proposal number with decreasing IoUs.}}
	\label{fig:iou-distribution}
	\vspace{-15pt}
\end{figure}

\subsection{The Use of High-quality Proposals}
RPN enhanced by the proposed guided anchoring scheme (GA-RPN) can generate much higher quality proposals than the conventional RPN.
We explore how to boost the performance of conventional two-stage detectors, through the use of such high-quality proposals.
Firstly, we study the IoU distribution of proposals generated by RPN and GA-RPN, as shown in Figure~\ref{fig:iou-distribution}.
There are two significant advantages of GA-RPN proposals over RPN proposals:
(1) the number of positive proposals is larger, and (2) the ratio of high-IoU
proposals is more significant.
A straight-forward idea is to replace RPN in existing models with the proposed GA-RPN
and train the model end-to-end.
However, this problem is non-trivial and adopting exactly the same settings as before can only bring limited gain (\eg, less than 1 point).
From our observation, the pre-requisite of using high-quality proposals is to adapt the distribution of training
samples in accordance to the proposal distribution. Consequently, we set a higher positive/negative threshold and use fewer samples when training
detectors end-to-end with GA-RPN compared to RPN.

Besides end-to-end training, we find that GA-RPN proposals are capable of boosting a trained
two-stage detector by a fine-tuning schedule. Specifically, given a trained model, we discard the proposal generation component, \eg, RPN, and use pre-computed GA-RPN proposals to finetune it for several epochs (3 epochs by default).
GA-RPN proposals are also used for inference.
This simple fine-tuning scheme can further improve the performance by a large margin,
with only a time cost of a few epochs.


\begin{table*}[!ht]
	\centering
	\caption{\small{Region proposal results on MS COCO.}}
	\small{
	\begin{tabular}{*{9}{c}}
		\toprule
		Method & Backbone & $\text{AR}_{100}$ & $\text{AR}_{300}$ & $\text{AR}_{1000}$ & $\text{AR}_{S}$ & $\text{AR}_{M}$ & $\text{AR}_{L}$ & runtime (s/img) \\
		\midrule
		SharpMask~\cite{SharpMask} & ResNet-50 & 36.4 & -   & 48.2 & 6.0  & 51.0 & 66.5 & 0.76 (unfair) \\
		GCN-NS~\cite{lu2018toward} & VGG-16 (SyncBN) & 31.6 & - & 60.7 & - & - & - & 0.10 \\
		AttractioNet~\cite{gidaris2016attendbmvc} & VGG-16 & 53.3 & -    & 66.2 & 31.5 & 62.2 & 77.7 & 4.00 \\
		ZIP~\cite{li2017zoom} & BN-inception & 53.9    & -    & 67.0    & 31.9 & 63.0 & 78.5 & 1.13 \\
		\midrule
		\multirow{3}{*}{RPN} & ResNet-50-FPN & 47.5 & 54.7 & 59.4 & 31.7 & 55.1 & 64.6 & \textbf{0.09} \\
		& ResNet-152-FPN & 51.9 & 58.0 & 62.0 & 36.3 & 59.8 & 68.1 & 0.16 \\
		& ResNeXt-101-FPN & 52.8 & 58.7 & 62.6 & 37.3 & 60.8 & 68.6 & 0.26 \\
		\midrule
		RPN+9 anchors & ResNet-50-FPN & 46.8 & 54.6 & 60.3 & 29.5 & 54.9 & 65.6 & 0.09 \\
		RPN+Focal Loss~\cite{lin2017focal} & ResNet-50-FPN & 50.2 & 56.6 & 60.9 & 33.9 & 58.2 & 67.5 & 0.09 \\
		RPN+Bounded IoU Loss~\cite{tychsen2018improving}  & ResNet-50-FPN & 48.3 & 55.1 & 59.6 & 33.0 & 56.0 & 64.3 & 0.09 \\
		RPN+Iterative & ResNet-50-FPN & 49.7 & 56.0 & 60.0 & 34.7 & 58.2 & 64.0 & 0.10 \\
		RefineRPN     & ResNet-50-FPN & 50.2 & 56.3 & 60.6 & 33.5 & 59.1 & 66.9 & 0.11 \\
		\midrule
		GA-RPN & ResNet-50-FPN & \textbf{59.2} & \textbf{65.2} & \textbf{68.5} & \textbf{40.9} & \textbf{67.8} & \textbf{79.0} & 0.13 \\
		\bottomrule
	\end{tabular}
	}
	\vspace{-10pt}
	\label{tab:rpn-results-coco}
\end{table*}

\section{Experiments}
\label{sec:experiments}

\subsection{Experimental Setting}

\noindent
\textbf{Dataset.}
We perform experiments on the challenging MS COCO 2017 benchmark~\cite{lin2014microsoft}.
We use the \emph{train} split for training and report the performance
on \emph{val} split. Detection results are reported on \emph{test-dev} split.

\noindent
\textbf{Implementation details.}
We use ResNet-50~\cite{he2016deep} with FPN~\cite{lin2017feature} as the
backbone network, if not otherwise specified.
As a common convention, we resize images to the scale of $1333\times800$,
without changing the aspect ratio.
We set $\sigma_1=0.2, \sigma_2=0.5$.
In the multi-task loss function, we simply use $\lambda_1=1, \lambda_2=0.1$ to
balance the location and shape prediction branches.
We use synchronized SGD over 8 GPUs with 2 images per GPU.
We train 12 epochs in total with an initial learning rate of 0.02,
and decrease the learning rate by 0.1 at epoch 8 and 11.
The runtime is measured on TITAN X GPUs.

\noindent
\textbf{Evaluation metrics.}
The results of RPN are measured with Average Recall (AR), which is the average
of recalls at different IoU thresholds (from 0.5 to 0.95). AR for 100, 300, and 1000 proposals
per image are denoted as $\text{AR}_{100}$, $\text{AR}_{300}$ and $\text{AR}_{1000}$.
The AR for small, medium, and large objects (AR$_S$, AR$_M$, AR$_L$)
are computed for 100 proposals.
Detection results are evaluated with the standard COCO metric, which averages mAP
of IoUs from 0.5 to 0.95.

\begin{table}[t]
	\centering
	\caption{\small{Detection results on MS COCO 2017 \emph{test-dev}.}}
	\addtolength{\tabcolsep}{-2pt}
	\small{
		\begin{tabular}{*{7}{c}}
			\toprule
			Method         & AP            & $\text{AP}_{50}$ & $\text{AP}_{75}$ & $\text{AP}_{S}$ & $\text{AP}_{M}$ & $\text{AP}_{L}$ \\
			\midrule
			Fast R-CNN     & 37.1          & 59.6             & 39.7             & 20.7            & 39.5            & 47.1            \\
			GA-Fast-RCNN   & \textbf{39.4} & 59.4             & 42.8             & 21.6            & 41.9            & 50.4            \\
			\midrule
			Faster R-CNN   & 37.1          & 59.1             & 40.1             & 21.3            & 39.8            & 46.5            \\
			GA-Faster-RCNN & \textbf{39.8} & 59.2             & 43.5             & 21.8            & 42.6            & 50.7            \\
			\midrule
			RetinaNet      & 35.9          & 55.4             & 38.8             & 19.4            & 38.9            & 46.5            \\
			GA-RetinaNet   & \textbf{37.1} & 56.9             & 40.0             & 20.1            & 40.1            & 48.0            \\
			\bottomrule
		\end{tabular}
	}
	\vspace{-10pt}
	\label{tab:det-results}
\end{table}

\begin{table}[h]
	\centering
	\caption{\small{Fine-tuning results on a trained Faster R-CNN.}}
	\addtolength{\tabcolsep}{-2pt}
	\small{
		\begin{tabular}{*{8}{c}}
			\toprule
			proposals & AP            & $\text{AP}_{50}$ & $\text{AP}_{75}$ & $\text{AP}_{S}$ & $\text{AP}_{M}$ & $\text{AP}_{L}$ \\
			\midrule
			-         & 37.4          & 58.9             & 40.3             & 20.8            & 41.1            & 49.5            \\
			RPN       & 37.3          & 58.6             & 40.1             & 20.4            & 40.6            & 49.8            \\
			GA-RPN    & \textbf{39.6} & \textbf{59.3}    & \textbf{43.0}    & \textbf{22.0}   & \textbf{42.8}   & \textbf{52.6}   \\
			\bottomrule
		\end{tabular}
	}
	\label{tab:finetune}
	\vspace{-10pt}
\end{table}

\subsection{Results}

We first evaluate our anchoring scheme by comparing the recall of GA-RPN
with the RPN baseline and previous state-of-the-art region proposal methods.
Meanwhile, we compare some variants of RPN.
``RPN+9 anchors'' denotes using 3 scales and 3 aspect ratios in each feature level,
while baselines use only 1 scale and 3 aspect ratios, following~\cite{lin2017feature}.
``RPN+Focal Loss'' and ``RPN+Bounded IoU Loss'' denotes adopting focal loss~\cite{lin2017focal} 
and bounded IoU Loss~\cite{tychsen2018improving} to RPN by substituting binary cross-entropy loss and smooth l1 loss, respectively.
``RPN+Iterative'' denotes applying two RPN heads consecutively,
with an additional $3\times3$ convolution between them.
``RefineRPN'' denotes a similar structure to~\cite{zhang2017single},
where anchors are regressed and classified twice with features before and after FPN.

\begin{figure}[t]
	\centering
	\includegraphics[width=\linewidth]{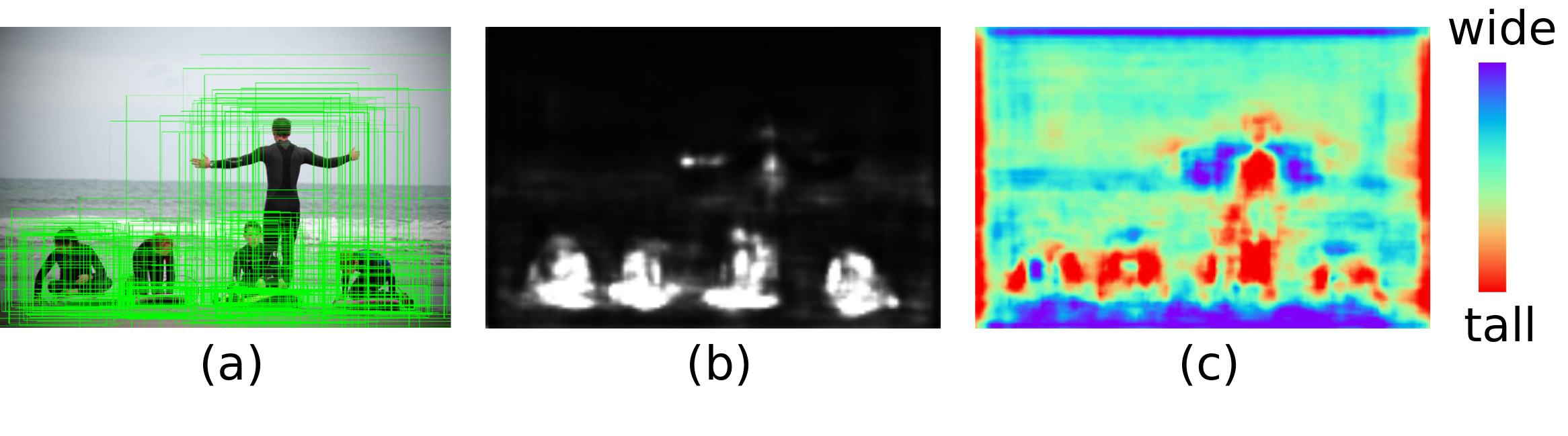}
	\vskip -0.25cm
	\caption{\small{Anchor prediction results. (a) input image and predict anchors;
		(b) predicted anchor location probability map; (c) predicted anchor aspect ratio.}}
	\label{fig:visualization}
	\vspace{-5pt}
\end{figure}

\begin{figure}[t]
	\centering
	\includegraphics[width=\linewidth]{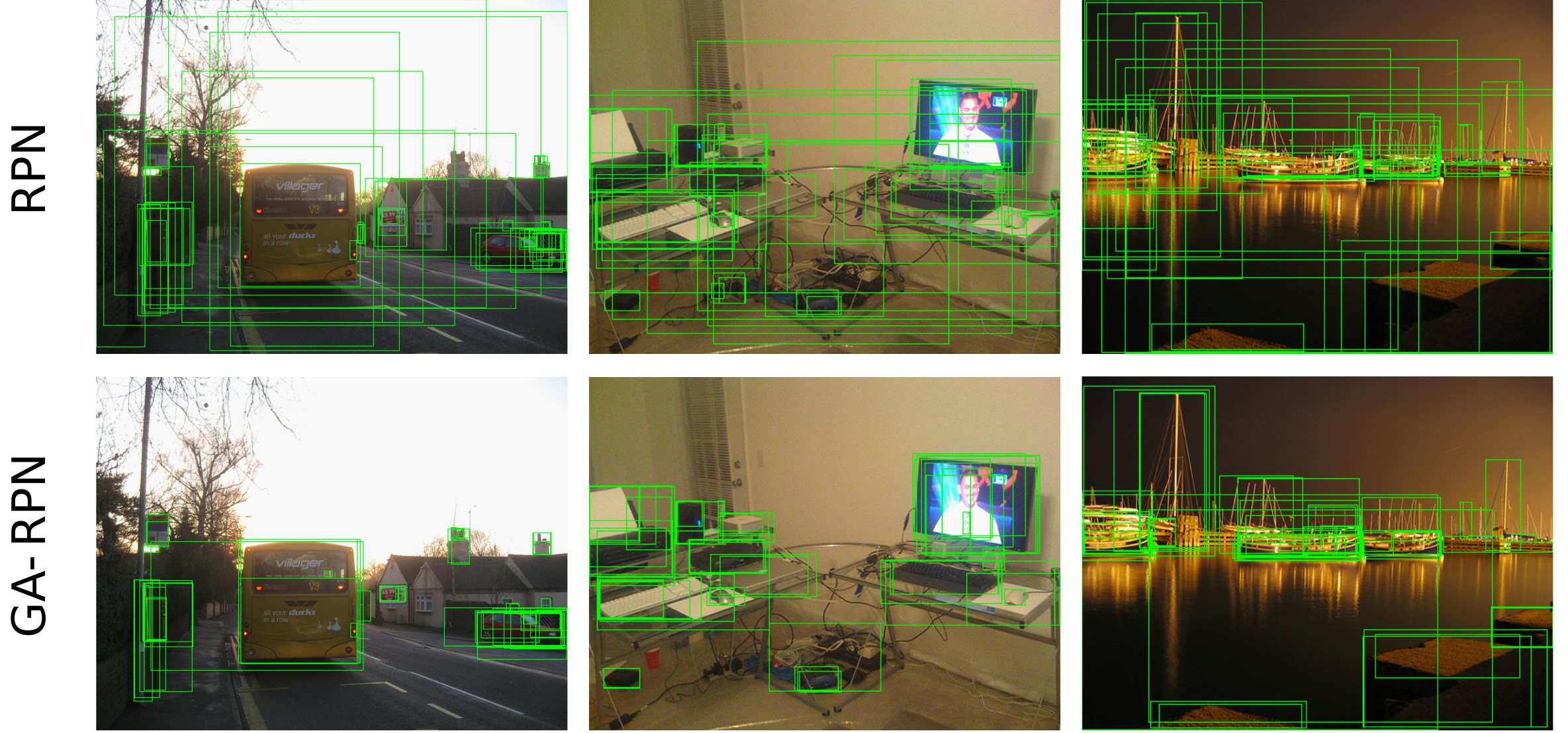}
	\caption{\small{Examples of RPN proposals (top row) and \algname~proposals (bottom row).}}
	\label{fig:examples}
	\vspace{-10pt}
\end{figure}

As shown in Table~\ref{tab:rpn-results-coco}, 
our method outperforms the RPN baseline by a large margin.
Specifically, it improves $\text{AR}_{300}$ by 10.5\% and $\text{AR}_{1000}$ by
9.1\% respectively.
Notably, GA-RPN with a small backbone can achieve a much higher recall than
RPN with larger backbones.
Our encouraging results are supported by the qualitative results shown in
Figure~\ref{fig:visualization}, where we show the sparse and arbitrary shaped
anchors and visualize the outputs of two branches.
It is observed that the anchors concentrate more on objects and provides
a good basis for the ensuing object proposal.
In Figure~\ref{fig:examples}, we show some examples of proposals generated upon
sliding window anchoring and guided anchoring.

Iterative regression and classification (``RPN+Iterative'' and ``RefineRPN'')
only brings limited gain to RPN,
which proves the importance of the aforementioned rule of alignment and consistency,
and simply refining anchors multiple times is not effective enough.
Keeping the center of anchors fixed and adapt features based on anchor shapes are crucial.

To investigate the generalization ability of guided anchoring and its power
to boost the detection performance, we integrate it into both two-stage and
single-stage detection pipelines, including Fast R-CNN~\cite{girshick2015fast},
Faster R-CNN~\cite{ren2015faster} and RetinaNet~\cite{lin2017focal}.
For two-stage detectors, we replace the original RPN with GA-RPN, and for
single-stage detectors, the sliding window anchoring scheme is replaced with
the proposed guided anchoring.
Results in Table~\ref{tab:det-results} show that guided anchoring not only
increases the proposal recall of RPN, but also improves the detection performance by a large margin.
With guided anchoring, the mAP of these detectors are improved by 2.3\%, 2.7\% and 1.2\% respectively.

To further study the effectiveness of high-quality proposals and investigate
the fine-tuning scheme,
we take a fully converged Faster R-CNN model and finetune it with pre-computed RPN or GA-RPN proposals.
We finetune the detector for 3 epochs, with the learning rate of 0.02, 0.002 and 0.0002 respectively.
The results are in Table~\ref{tab:finetune} illustrate that RPN proposals cannot
bring any gain,
while the high-quality GA-RPN proposals bring 2.2\% mAP improvement to the
trained model with only a time cost of 3 epochs.

\begin{table}[t]
	\centering
	\caption{\small{The effects of each module in our design. L., S., and F.A. denote location, shape, and feature adaptation, respectively.}}
		\addtolength{\tabcolsep}{-2pt}
	\small{
		\begin{tabular}{*{9}{c}}
			\toprule
			L.   & S.      & F.A. & $\text{AR}_{100}$ & $\text{AR}_{300}$ & $\text{AR}_{1000}$ & $\text{AR}_{S}$ & $\text{AR}_{M}$ & $\text{AR}_{L}$ \\
			\midrule
			           &            &                  & 47.5              & 54.7              & 59.4               & 31.7            & 55.1            & 64.6            \\
			\checkmark &            &                  & 48.0              & 54.8              & 59.5               & 32.3            & 55.6            & 64.8            \\
			           & \checkmark &                  & 53.8              & 59.9              & 63.6               & 36.4            & 62.9            & 71.7            \\
			\checkmark & \checkmark &                  & 54.0              & 60.1              & 63.8               & 36.7            & 63.1            & 71.5            \\
			\checkmark & \checkmark & \checkmark       & \textbf{59.2}     & \textbf{65.2}     & \textbf{68.5}      & \textbf{40.9}   & \textbf{67.8}   & \textbf{79.0}   \\
			\bottomrule
		\end{tabular}
	}
	\vspace{-5pt}
	\label{tab:components}
\end{table}

\begin{table}[t]
	\centering
	\caption{\small{Results of different location threshold $\epsilon_L$.}}
	\addtolength{\tabcolsep}{-2pt}
	\small{
		\begin{tabular}{*{9}{c}}
			\toprule
			$\epsilon_L$ & \#anchors/image & $\text{AR}_{100}$ & $\text{AR}_{300}$ & $\text{AR}_{1000}$ & fps \\
			\midrule
			0         & 75583 (100.0\%) & 59.2              & 65.2              & 68.5               & 7.8 \\
			0.01      & 22274 (29.4\%)  & 59.2              & 65.2              & 68.5               & 8.0 \\
			0.05      & 5251 (6.5\%)    & 59.1              & 65.1              & 68.2               & 8.2 \\
			0.1       & 2375 (3.2\%)    & 59.0              & 64.7              & 67.2               & 8.2 \\
			\bottomrule
		\end{tabular}
	}
	\vspace{-5pt}
	\label{tab:location-thr}
\end{table}

\subsection{Ablation Study}

\noindent
\textbf{Model design.}
We omit different components in our design to investigate the effectiveness
of each component, including location prediction, shape prediction and
feature adaption.
Results are shown in Table~\ref{tab:components}. The shape prediction branch
is shown effective which leads to a gain of $4.2\%$.
The location prediction branch brings marginal improvement. Nevertheless, the importance of this branch is reflected in its usefulness of obtaining sparse anchors leading to more efficient inference.
The obvious gain brought by the feature adaption module suggests the necessity
of rearranging the feature map according to predicted anchor shapes.
This module helps to capture information corresponding to anchor scopes,
especially for large objects.

\begin{figure}[tb]
	\centering
	\begin{subfigure}{.23\textwidth}
		\centering
		\includegraphics[width=\textwidth]{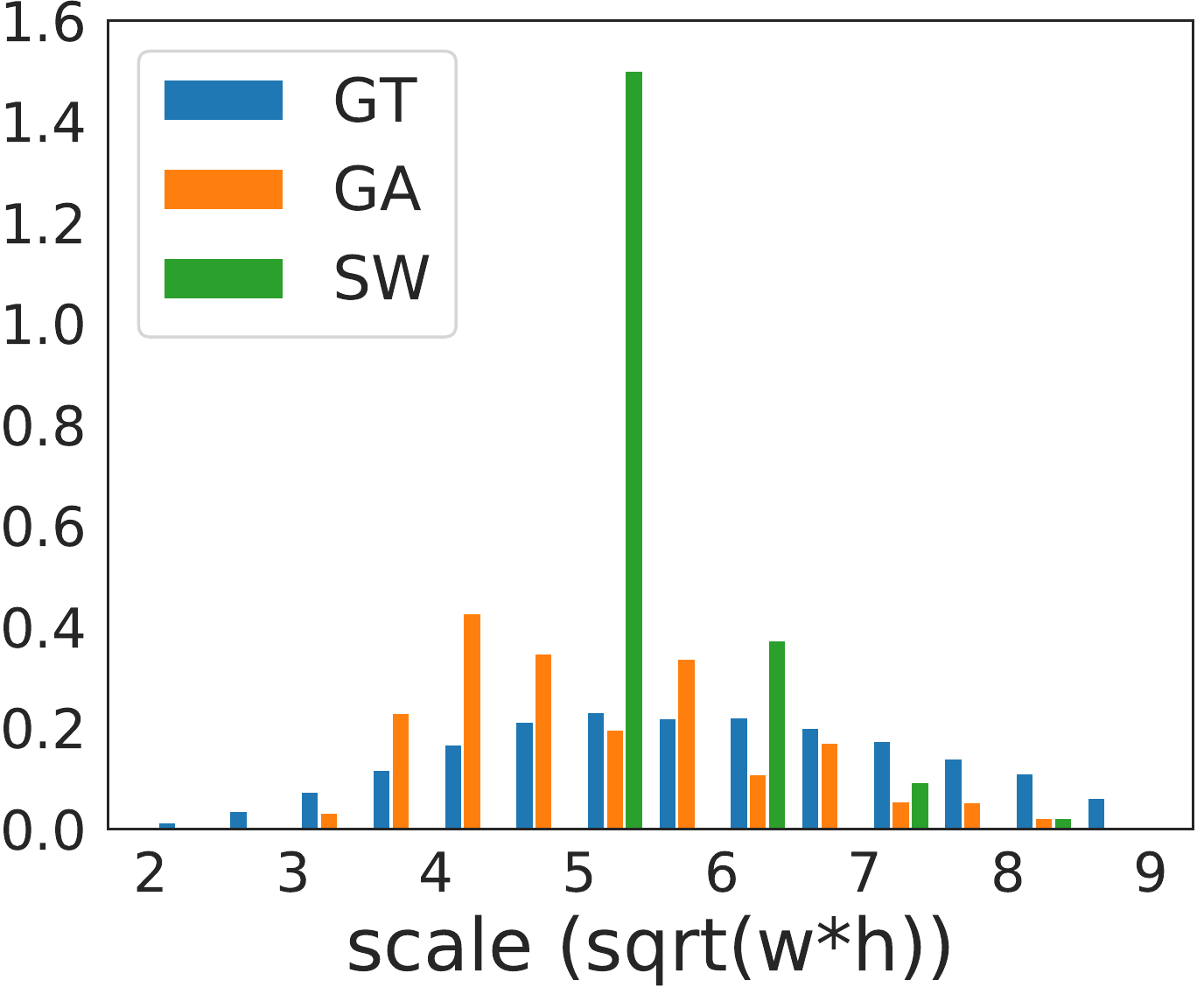}
		\caption{}
		\label{subfig:anchor-scale}
	\end{subfigure}
	\vspace{-0.2cm}
	\begin{subfigure}{.23\textwidth}
		\centering
		\includegraphics[width=\textwidth]{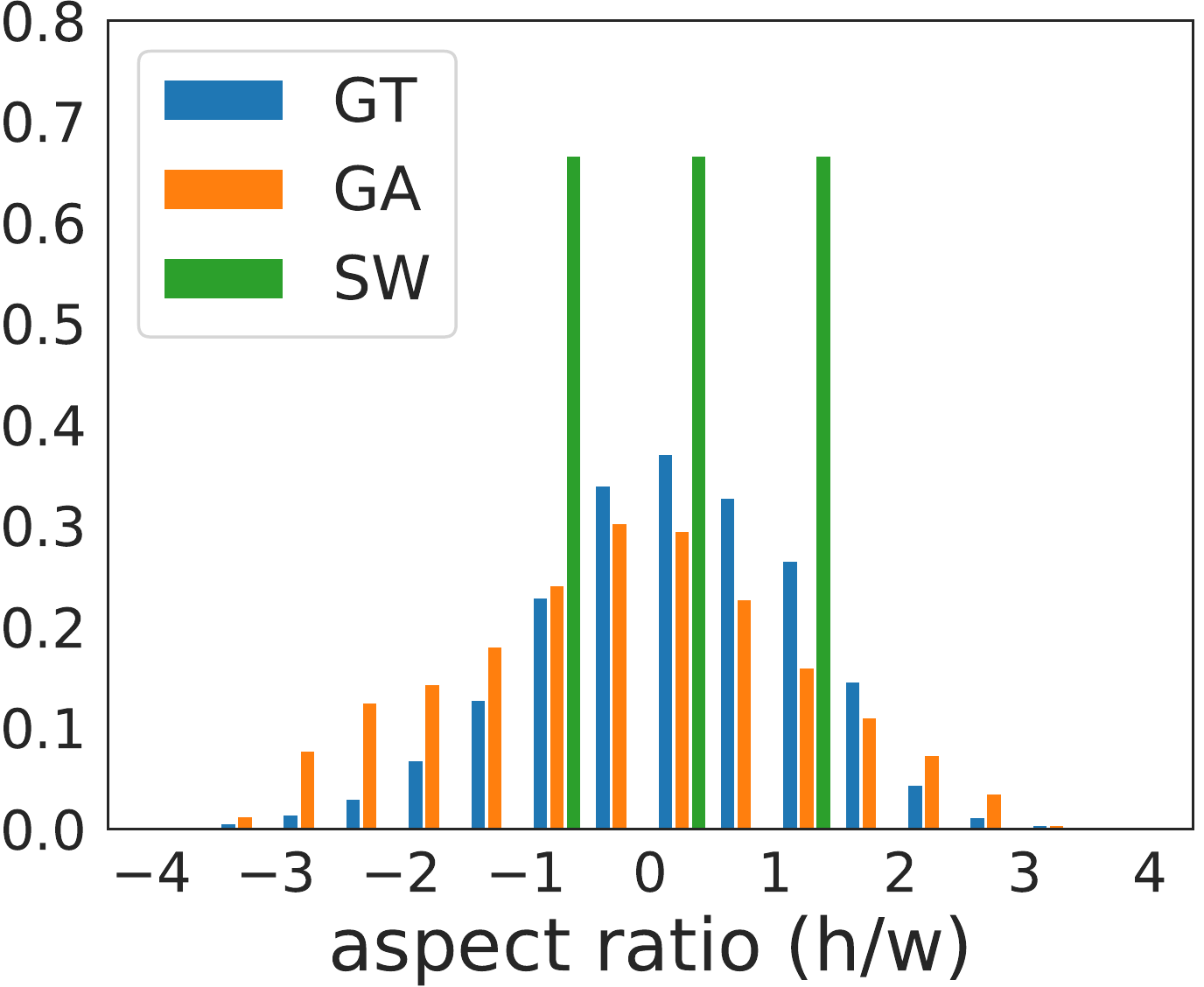}
		\caption{}
		\label{subfig:anchor-ar}
	\end{subfigure}
	\vspace{-0.2cm}
	\caption{\small{(a) Anchor scale and (b) aspect ratio distributions of different anchoring schemes.
		The x-axis is reduced to log-space by apply $\log_2(\cdot)$ operator. GT, GA, SW indicates ground truth, guided anchoring, sliding window, respectively.}}
	\label{fig:anchor-dist}
	\vspace{-0.6cm}
\end{figure}

\noindent
\textbf{Anchor location.}
The location threshold $\epsilon_L$ controls the sparsity of anchor distribution.
Adopting different thresholds will yield different numbers of anchors.
To reveal the influence of $\epsilon_L$ on efficiency and performance,
we vary the threshold and compare the following results: the average number
of anchors per image, recall of final proposals and the inference runtime.
From Table~\ref{tab:location-thr} we can observe that the objectness scores of
most background regions are close to 0, so a small $\epsilon_L$ can
greatly reduce the number of anchors by more than 90\%, with only a minor
decrease on recall rate.
It is noteworthy that the head in RPN is just one convolutional layer,
so the speedup is not apparent. Nevertheless, a significant reduction in the number of anchors offers a possibility to perform more efficient inference with a heavier head.

\noindent
\textbf{Anchor shape.}
We compare the set of generated anchors of our method with sliding window
anchors of pre-defined shapes.
Since our method predicts only one anchor at each location of the feature map
instead of $k$ ($k=3$ in our baseline) anchors of different scales and
aspect ratios, the total anchor number is reduced by $\frac{1}{k}$.
We present the scale and aspect ratio distribution of our anchors with
sliding window anchors in Figure~\ref{fig:anchor-dist}.
The results show great advantages of the guided anchoring scheme over
predefined anchor scales and shapes. The predicted anchors cover a much wider
range of scales and aspect ratios, which have a similar distribution to
ground truth objects and provide a pool of initial anchors with higher
coverage on objects.

\noindent
\textbf{Feature adaption.}
The feature adaption module improves the recall by a large margin,
proving that a remedy of features consistency is essential.
We claim that the improvement not only comes from adopting deformable
convolution, but also results from our design of using anchor shape predictions
to predict the offset of the deformable convolution layer.
If we simply add a deformable convolution layer after anchor generation,
the results of AR100/AR300/AR1000 are 56.1\%/62.4\%/66.1\%, which are inferior to
results from our design.

\noindent
\textbf{Alignment and consistency rule.}
We verify the necessity of the two proposed rules.
The alignment rule suggests that we should keep the anchor centers aligned with
feature map pixels.
According to the consistency rule, we design the feature adaption module to
refine the features.
Results in Table~\ref{tab:alignment_consistency} show the importance of these rules.
1) From row 1 and 2, or row 3 and 4, we learn that predicting both the shape
and center offset instead of just predicting the shape harms the performance.
2) The comparison between row 1 and 3, or row 2 and 4 shows the impact of consistency.

\begin{table}[t]
	\centering
	\caption{\small{The effects of alignment and consistency rules. C.A. and F.A. denote center alignment (alignment rule) and feature adaption (consistency rule) respectively.}}
	\addtolength{\tabcolsep}{-2pt}
	\small{
		\begin{tabular}{*{9}{c}}
			\toprule
			C.A.   & F.A.      & $\text{AR}_{100}$ & $\text{AR}_{300}$ & $\text{AR}_{1000}$ & $\text{AR}_{S}$ & $\text{AR}_{M}$ & $\text{AR}_{L}$ \\
			\midrule
			&          &              51.7              & 58.0              &  61.6              & 33.8            & 60.9            & 70.0         \\
			\checkmark &    &         54.0              & 60.1              & 63.8               & 36.7            & 63.1            & 71.5        \\
			& \checkmark &            57.2              & 63.6              & 66.8               & 38.3            & 66.1            & 77.8        \\
			\checkmark & \checkmark &  \textbf{59.2}     & \textbf{65.2}     & \textbf{68.5}      & \textbf{40.9}   & \textbf{67.8}   & \textbf{79.0}   \\
			\bottomrule
		\end{tabular}
	}
	\vspace{-0.2cm}
	\label{tab:alignment_consistency}
\end{table}

\begin{table}[t]
	\centering
	\caption{\small{Exploration of utilizing high-quality proposals.}}
	\addtolength{\tabcolsep}{-2pt}
	\small{
		\begin{tabular}{*{6}{c}}
			\toprule
			proposal                & num  & IoU thr & AP            & $\text{AP}_{50}$ & $\text{AP}_{75}$ \\
			\midrule
			\multirow{4}{*}{RPN}    & 1000 & 0.5         & 36.7          & 58.8             & 39.3             \\
			& 1000 & 0.6         & 37.2          & 57.1             & 40.5             \\
			& 300  & 0.5         & 36.1          & 57.6             & 39.0             \\
			& 300  & 0.6         & 37.0          & 56.3             & 39.5             \\
			\midrule
			\multirow{4}{*}{GA-RPN} & 1000 & 0.5         & 37.4          & \textbf{59.9}    & 40.0             \\
			& 1000 & 0.6         & 38.9          & 59.0             & 42.4             \\
			& 300  & 0.5         & 37.5          & 59.6             & 40.4             \\
			& 300  & 0.6         & \textbf{39.4} & 59.3             & \textbf{43.2}    \\
			\bottomrule
		\end{tabular}
	}
	\label{tab:proposal-utilization}
	\vspace{-10pt}
\end{table}

\noindent
\textbf{The use of high-quality proposals.}
Despite with high-quality proposals, training a good detector remains a non-trivial problem.
As illustrated in Figure~\ref{fig:iou-distribution}, GA-RPN proposals provide
more candidates of high IoU. This suggests that we can use fewer proposals
for training detectors.
We test different numbers of proposals and different IoU thresholds to assign
labels for foreground/background on Fast R-CNN.

From the results in Table~\ref{tab:proposal-utilization}, we observe that:
(1) Larger IoU threshold is important for taking advantage of high-quality proposals.
By focusing on positive samples of higher IoU, there will be fewer false
positives and the features for classification are more discriminative.
Since we assign negative labels to proposals with IoU less than 0.6 during training,
$\text{AP}_{0.5}$ will decrease while AP of high IoUs will
increase by a large margin, and the overall AP is much higher.
(2) Using fewer proposals during training and testing can benefit the learning
if the recall is high enough.
Fewer proposals lead to a lower recall, but will simplify the learning process,
since there are more hard samples in low-score proposals.
When training with RPN proposals, the performance will decrease if we use only
300 proposals, because the recall is not sufficient and many objects get missed.
However, GA-RPN guarantees high recall even with fewer proposals, thus training
with 300 proposals could still boost the final mAP.

\noindent
\textbf{Hyper-parameters.}
Our method is insensitive to hyper-parameters.
\textbf{(1)}
As we sample $3$, $9$, $15$ pairs to approximate Eq.\eqref{eq:shape_approx}, we respectively obtain AR@1000 $68.3\%$, $68.5\%$, $68.5\%$.
\textbf{(2)}
We set $\lambda_2=0.1$ to balance the loss terms by default.
We obtain $68.4\%$ with $\lambda_2 = 0.2 \text{ or } 0.05$ and
$68.3\%$ with $\lambda_2 = 0.02$.
\textbf{(3)}
We vary $\sigma_1$ within $[0.1, 0.5]$ and $\sigma_2$ within $[0.2, 1.0]$,
and the performance remains comparable (between $68.1\%$ and $68.5\%$).


\section{Conclusion}
\label{sec:conclusion}

We have proposed the Guided Anchoring scheme, which leverages semantic features
to guide the anchoring. It generates non-uniform anchors of arbitrary shapes by
jointly predicting the locations and anchor shapes dependent on locations.
The proposed method achieves 9.1\% higher recall with 90\% fewer anchors than
the RPN baseline using the sliding window scheme. It can also be applied to
various anchor-based detectors to improve the performance by as much as 2.7\%.

\vspace{-10pt}

\paragraph{Acknowledgment}
This work is partially supported by the Collaborative Research grant from SenseTime Group (CUHK Agreement No. TS1610626 \& No. TS1712093), the General Research Fund (GRF) of Hong Kong (No. 14236516, No. 14203518 \& No. 14224316), and Singapore MOE AcRF Tier 1 (M4012082.020).


\appendix

\begin{figure*}[h]
	\begin{center}
		\includegraphics[width=.95\linewidth]{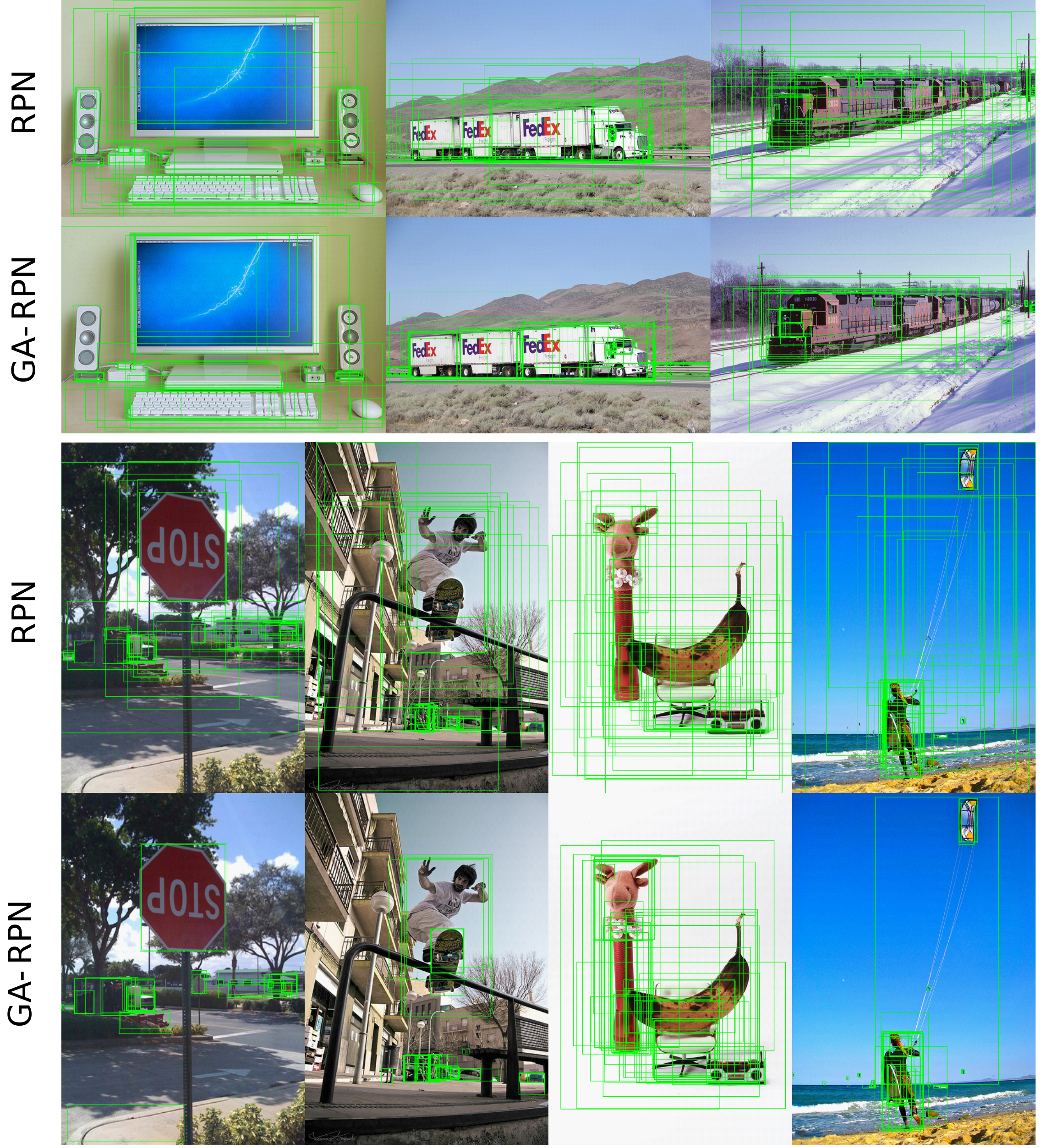}
		\label{fig:examples1}
	\end{center}
	\vspace{3cm}
\end{figure*}

\begin{figure*}[h]
	\begin{center}
		\includegraphics[width=.95\linewidth]{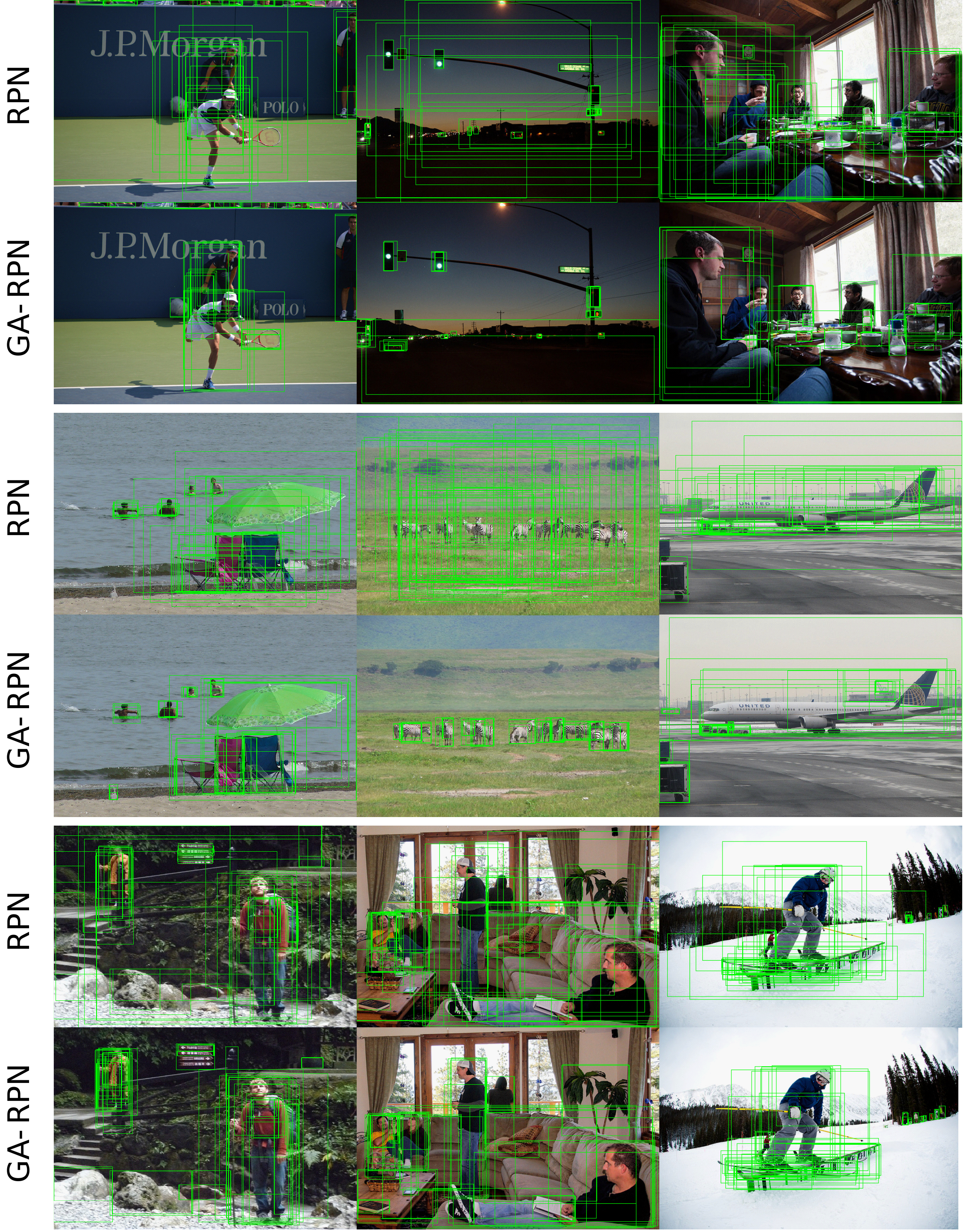}
		\caption{Examples of RPN proposals (top row) and \algname~proposals (bottom row).}
		\label{fig:examples2}
	\end{center}%
\end{figure*}

{\small
	\bibliographystyle{ieee_fullname}
	\bibliography{sections/egbib}
}

\end{document}